\documentclass[letterpaper, 10 pt, conference, dvipsnames]{ieeeconf} 
\IEEEoverridecommandlockouts
\newcommand\copyrighttext{%
	\footnotesize This work has been submitted to the IEEE for possible publication. Copyright may be transferred without notice, after which this version may no longer be accessible.%
}
\newcommand\copyrightnotice{%
	\begin{tikzpicture}[remember picture,overlay]
	\node[anchor=south,yshift=10pt, xshift=10pt] at (current page.south) {\fbox{\parbox{\dimexpr\textwidth-\fboxsep-\fboxrule\relax}{\copyrighttext}}};
	\end{tikzpicture}%
}

\usepackage{printlen}

\usepackage[T1]{fontenc}
\usepackage{amsmath,amsfonts, amssymb}
\usepackage{cite}
\usepackage{stfloats}
\usepackage{graphicx}
\usepackage[caption=false,font=scriptsize]{subfig}
\usepackage{url}
\usepackage{hyperref}

\usepackage[per-mode = power]{siunitx}
\usepackage{acronym}
\usepackage[none]{hyphenat}
\usepackage{glossaries}
\usepackage[switch]{lineno}

\usepackage{bbm} 

\usepackage{algorithm}
\usepackage{algpseudocode}
\usepackage{minted}

\usepackage{booktabs}
\usepackage{tabularx}
\usepackage{multirow}
\usepackage{pifont}

\usepackage{wheelchart}
\usepackage{etoolbox}
\usepackage{tikzsymbols}

\usepackage{tikz, pgfplots, pgfplotstable}
\usetikzlibrary{positioning, arrows.meta, fit, backgrounds, calc, shapes.geometric, decorations.markings, patterns, decorations.pathreplacing, fadings, shadows, decorations.pathreplacing, calligraphy}
\usepgfplotslibrary{statistics}
\usepgfplotslibrary{groupplots}
\pgfplotsset{compat=1.18}

\usepackage{xcolor}
\definecolor{Black}{HTML}{000000}
\definecolor{Blue}{HTML}{0065bd}
\definecolor{Bluelight}{HTML}{D6E8F7}

\definecolor{Bluestrong}{HTML}{003359}
\definecolor{Red}{HTML}{8C000F}

\definecolor{OrangePP}{HTML}{E97132}
\definecolor{Green}{HTML}{A2AD00}
\definecolor{GreenCR}{HTML}{008000}
\definecolor{LightGray}{HTML}{e7e7e7}
\definecolor{Gray}{HTML}{7f7f7f}
\definecolor{Gray-opac}{HTML}{d8d8d8}
\definecolor{MyDarkBlue}{RGB}{14,40,65}
\definecolor{Red}{RGB}{196,7,27}

    \definecolor{TUMBlue}{HTML}{0065BD}
    \definecolor{TUMBlack}{HTML}{000000}
    \definecolor{TUMWhite}{HTML}{FFFFFF}
    \definecolor{TUMDarkBlue}{HTML}{005293}
    \definecolor{TUMLightBlue}{HTML}{64A0C8}
    \definecolor{TUMLighterBlue}{HTML}{98C6EA}
    \definecolor{TUMGray}{HTML}{999999}
    \definecolor{TUMOrange}{HTML}{E37222}
    \definecolor{TUMGreen}{HTML}{A2AD00}
    \definecolor{TUMLightGray}{HTML}{DAD7CB}

    \definecolor{TUMBlueBrand}{HTML}{3070B3}
    \definecolor{TUMBlueDark}{HTML}{072140}
    \definecolor{TUMBlueDark1}{HTML}{0A2D57}
    \definecolor{TUMBlueDark2}{HTML}{0E396E}
    \definecolor{TUMBlueDark3}{HTML}{114584}
    \definecolor{TUMBlueDark4}{HTML}{14519A}
    \definecolor{TUMBlueDark5}{HTML}{165DB1}
    \definecolor{TUMBlueLight}{HTML}{5E94D4}
    \definecolor{TUMBlueLightDark}{HTML}{9ABCE4}
    \definecolor{TUMBlueLight2}{HTML}{C2D7EF}
    \definecolor{TUMBlueLight3}{HTML}{D7E4F4}
    \definecolor{TUMBlueLight4}{HTML}{E3EEFA}
    \definecolor{TUMBlueLight5}{HTML}{F0F5FA}
    
    \definecolor{TUMYellow}{HTML}{FED702}
    \definecolor{TUMYellowDark}{HTML}{CBAB01}
    \definecolor{TUMYellow1}{HTML}{FEDE34}
    \definecolor{TUMYellow2}{HTML}{FEE667}
    \definecolor{TUMYellow3}{HTML}{FEEE9A}
    \definecolor{TUMYellow4}{HTML}{FEF6CD}
    
    \definecolor{TUMWebOrange}{HTML}{F7811E}
    \definecolor{TUMOrangeDark}{HTML}{D99208}
    \definecolor{TUMOrange1}{HTML}{F9BF4E}
    \definecolor{TUMOrange2}{HTML}{FAD080}
    \definecolor{TUMOrange3}{HTML}{FCE2B0}
    \definecolor{TUMOrange4}{HTML}{FEF4E1}
    
    \definecolor{TUMPink}{HTML}{B55CA5}
    \definecolor{TUMPinkDark}{HTML}{9B468D}
    \definecolor{TUMPink1}{HTML}{C680BB}
    \definecolor{TUMPink2}{HTML}{D6A4CE}
    \definecolor{TUMPink3}{HTML}{E6C7E1}
    \definecolor{TUMPink4}{HTML}{F6EAF4}
    
    \definecolor{TUMBlueBright}{HTML}{8F81EA}
    \definecolor{TUMBlueBrightDark}{HTML}{6955E2}
    \definecolor{TUMBlueBright1}{HTML}{B6ACF1}
    \definecolor{TUMBlueBright2}{HTML}{C9C2F5}
    \definecolor{TUMBlueBright3}{HTML}{DCD8F9}
    \definecolor{TUMBlueBright4}{HTML}{EFEDFC}
    
    \definecolor{TUMRed}{HTML}{EA7237}
    \definecolor{TUMRedDark}{HTML}{D95117}
    \definecolor{TUMRed1}{HTML}{EF9067}
    \definecolor{TUMRed2}{HTML}{F3B295}
    \definecolor{TUMRed3}{HTML}{F6C2AC}
    \definecolor{TUMRed4}{HTML}{FBEADA}
    
    \definecolor{TUMWebGreen}{HTML}{9FBA36}
    \definecolor{TUMGreenDark}{HTML}{7D922A}
    \definecolor{TUMGreen1}{HTML}{B6CE55}
    \definecolor{TUMGreen2}{HTML}{C7D97D}
    \definecolor{TUMGreen3}{HTML}{D8E5A4}
    \definecolor{TUMGreen4}{HTML}{E9F1CB}
    
    \definecolor{TUMGrey1}{HTML}{20252A}
    \definecolor{TUMGrey2}{HTML}{333A41}
    \definecolor{TUMGrey3}{HTML}{475058}
    \definecolor{TUMGrey4}{HTML}{6A757E}
    \definecolor{TUMGrey7}{HTML}{DDE2E6}
    \definecolor{TUMGrey8}{HTML}{EBECEF}
    \definecolor{TUMGrey9}{HTML}{FBF9FA}
    \definecolor{TUMWebWhite}{HTML}{FFFFFF}

\newcommand{\blob}[1]{%
    \tikz\fill[#1] (0,0) circle (2mm);
}

\graphicspath{{./figures/}}

\title{\LARGE \bf
Disengagement Analysis and Field Tests of a Prototypical Open-Source Level 4 Autonomous Driving System}

\author{Marvin Seegert, Christian Oefinger, Korbinian Moller, Christoph Bank, Johannes Betz%
\thanks{M. Seegert, C. Oefinger K. Moller, C. Bank, and J. Betz are with the Professorship of Autonomous Vehicle Systems, TUM School of Engineering and Design, Technical University of Munich, 85748 Garching, Germany; Munich Institute of Robotics and Machine Intelligence (MIRMI).}%
}

\newacronym{av}{AV}{Autonomous Vehicle}%
\newacronym{ad}{AD}{Autonomous Driving}%
\newacronym{ads}{ADS}{Autonomous Driving System}%
\newacronym{aw}{AW}{Autoware}%
\newacronym{adas}{ADAS}{Advanced Driver Assistance System}%
\newacronym{ros2}{ROS~2}{Robot Operating System 2}%
\newacronym{sd}{SD}{Safety Driver}
\newacronym{de}{disengagement}{disengagement}
\newacronym{cadmv}{CA DMV}{California Department of Motor Vehicles}
\newacronym{tod}{ToD}{Teleoperated Driving}
\newacronym{ttc}{TTC}{Time to Collision}
\newacronym{pet}{PET}{Post Encroachment Time}

\begin{document}
\bstctlcite{BSTcontrol}

\maketitle
\copyrightnotice%
\vspace{-3.8mm}

\begin{figure*}[b!]
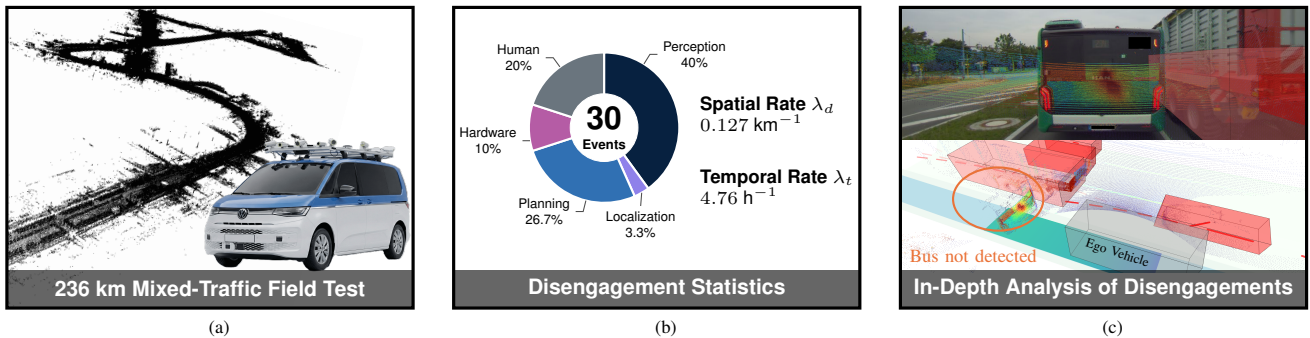

    \vspace{-4.0 mm}
    \centering
    \subfloat[]{%
        \begin{minipage}{0.325\textwidth}
            \centering
            %
    \input{figures/pipeline1}%

        \end{minipage}
        \label{fig:edgar}
    }
    \hfill
    \subfloat[]{%
        \begin{minipage}{0.325\textwidth}
            \centering
            %
    \begin{tikzpicture}
    \def\panelW{5.4cm} 
    \def\panelH{4.0cm}

    \useasboundingbox (0,0) rectangle (\panelW, \panelH);
    
    \begin{scope}
        \clip (0,0) rectangle (\panelW, \panelH);

        \begin{scope}[shift={(2.0, 2.4)}]
            
            \def\R{1.0} 
            
            \draw[fill=TUMBlueDark, draw=white, thick] 
              (0,0) -- (90:\R) arc (90:-54:\R) -- cycle;
              
            \draw[fill=TUMBlueBright, draw=white, thick] 
              (0,0) -- (-54:\R) arc (-54:-66:\R) -- cycle;

            \draw[fill=TUMBlueBrand, draw=white, thick] 
              (0,0) -- (-66:\R) arc (-66:-162:\R) -- cycle;

            \draw[fill=TUMPink, draw=white, thick] 
              (0,0) -- (-162:\R) arc (-162:-198:\R) -- cycle;

            \draw[fill=TUMGrey4, draw=white, thick] 
              (0,0) -- (-198:\R) arc (-198:-270:\R) -- cycle;

            \draw[fill=white, draw=white] (0,0) circle (\R*0.45);
            
            \node[align=center, font=\sffamily\bfseries] at (0,0) {\large 30\\[-1.5mm]\tiny Events};

            \draw[thin, draw=black!70] (58:\R) -- (58:\R+0.15) -- ++(0.1,0) 
                node[right, font=\sffamily\tiny, align=center, inner sep=2pt] {Perception\\40\%};
            \draw[thin, draw=black!70] (-60:\R) -- (-65:\R+0.15) 
                node[below, font=\sffamily\tiny,align=center, inner sep=2pt] {Localization\\3.3\%};
            \draw[thin, draw=black!70] (-104:\R) -- (-104:\R+0.15) -- ++(-0.1,0) 
                node[left, font=\sffamily\tiny, align=center, inner sep=2pt] {Planning\\26.7\%};
            \draw[thin, draw=black!70] (-170:\R) -- (-170:\R+0.0) -- ++(-0.1,0) 
                node[left, font=\sffamily\tiny,align=center, inner sep=2pt] {Hardware\\10\%};
            \draw[thin, draw=black!70] (-234:\R) -- (-234:\R+0.15) -- ++(-0.1,0) 
                node[left, font=\sffamily\tiny, align=center, inner sep=2pt] {Human\\20\%};

        \end{scope}

        \node[anchor=center, align=left, font=\sffamily\scriptsize] at (4.3, 2.1) {
            \textbf{Spatial Rate $\lambda_d$}\\
            $0.127\text{ km}^{-1}$\\[4mm]
            \textbf{Temporal Rate $\lambda_t$}\\
            $4.76\text{ h}^{-1}$
        };

        \node[
            fill=black, 
            fill opacity=0.6, 
            text opacity=1, 
            text=white, 
            font=\sffamily\bfseries\footnotesize,
            minimum width=\panelW, 
            minimum height=0.4cm, 
            anchor=south
        ] at (\panelW/2, 0) {Disengagement Statistics};

    \end{scope}

    \draw[black, very thick] (0,0) rectangle (\panelW, \panelH);
    
\end{tikzpicture}%

        \end{minipage}
    }
    \hfill
    \subfloat[]{%
        \begin{minipage}{0.325\textwidth}
            \centering
            %
    \input{figures/pipeline3}%

        \end{minipage}
        \label{fig:perception_scenario}
    }
    \vspace{-3mm}
    \caption{Overview of the real-world evaluation of an open-source Level 4 \acrlong{ads}. (a) A prototypical research vehicle running the Autoware Universe stack was deployed across \SI{236}{km} of mixed-traffic environments. (b) Quantitative analysis of 30 disengagements revealed a spatial rate of \SI{0.127}{\per\kilo\meter}, with perception (40\%) and planning (26.7\%) limitations as the primary bottlenecks. (c) In-depth analysis exposes critical real-world situations, such as the perception system failing to detect a bus, demonstrating that rare anomalies can aggregate into system-level failures.}
    \label{fig:title_figure}
\end{figure*}%


\begin{abstract}
Proprietary Autonomous Driving Systems are typically evaluated through disengagements, unplanned manual interventions to alter vehicle behavior, as annually reported by the \acrlong{cadmv}. However, the real-world capabilities of prototypical open-source Level 4 vehicles over substantial distances remain largely unexplored. This study evaluates a research vehicle running an Autoware-based software stack across 236 km of mixed traffic. By classifying 30 disengagements across 26 rides with a novel five-level criticality framework, we observed a spatial disengagement rate of 0.127 $\text{km}^{\text{-1}}$. Interventions predominantly occurred at lower speeds near static objects and traffic lights. Perception and Planning failures accounted for 40\% and 26.7\% of disengagements, respectively, largely due to object-tracking losses and operational deadlocks caused by parked vehicles. Frequent, unnecessary interventions highlighted a lack of trust on the part of the safety driver. These results show that while open-source software enables extensive operations, disengagement analysis is vital for uncovering robustness issues missed by standard metrics.
\end{abstract}


\section{Introduction}
\label{sec:introduction}

The evolution of \gls{ad} has been largely defined by two parallel paths: proprietary systems documented in regulatory reports and open-source platforms that offer transparency but limited real-world validation. While the \gls{cadmv} disengagement reports provide data for some commercial entities, they are often criticized for ambiguous definitions and lack of technical detail \cite{Favaro2018, KoopmanInterview2019, Skokan2025}. Conversely, open-source stacks like Autoware~\cite{Kato2018} have broadened access to \gls{ad} research. However, existing literature on their deployment remains confined to controlled environments, low-speed campus laps, or singular test runs~\cite{Korzelius2025, Zhang2025, Ganesan2025}, leaving the Level~4 capabilities of prototypical open-source systems in mixed traffic over substantial distances largely unexplored.

Evaluating these systems is challenging due to "long-tail" edge cases that accumulate into failures over extended operation~\cite{Liu2024}. High accuracy in standard dataset-derived metrics fails to capture how marginal error rates in dynamic environments require frequent safety-relevant interventions. This is further complicated by safety drivers who initiate unnecessary interventions due to perceived unreliability. These false positive events underscore the need for a framework that distinguishes genuine software limitations from interventions driven by a lack of trust.

To address these gaps, this study evaluates an adapted version of the Autoware Universe~\cite{AWrepo} stack through~$26$ real-world rides covering \SI{236}{\kilo\meter} of mixed traffic. A five-level criticality framework is introduced that supplements disengagement counts by categorizing events by severity and operational impact. This framework classifies disengagements, ranging from critical safety failures to operational deadlocks, allowing a distinction between software limitations and precautionary driver intervention.

By examining failure modes like tracking losses and deadlocks, this study reveals how infrequent anomalies aggregate into system-level failures. These results underscore the importance of domain-specific, real-world validation to catch failures missed by offline metrics. Ultimately, this work provides a benchmark for prototypical open-source \gls{ad} systems and outlines actionable improvements to enhance situational awareness, including complex, long-tail scenarios.

\section{Related Work}
\label{sec:relatedwork}

A substantial body of literature~\cite{Wang2020, Favaro2018, Kohanpour2025, Feng2020, Boggs2020, Ward2024, Lee2025, Wang2019, Ansarinejad2025, Ali2025, Lv2018} relies on the annual disengagement reports published by the \gls{cadmv} to evaluate autonomous driving performance~\cite{CCRdisengagement}, spanning data from~2014~\cite{Dixit2016} to~2024~\cite{Skokan2025}. These reports provide a broad overview of commercial vehicle testing, with some studies observing $2\times 10^{-4}$ up to $3$ disengagements per mile~\cite{Wang2020}. However, relying exclusively on \gls{cadmv} data is criticized for inconsistent reporting~\cite{Favaro2018, KoopmanInterview2019} and for the ambiguous regulatory definition of a disengagement~\cite{Skokan2025}.

To extract deeper insights, several studies propose frameworks for categorizing the root causes of \gls{cadmv} disengagements~\cite{Kohanpour2025}. For instance,~\cite{Boggs2020} maps failures to specific autonomous-driving modules, while~\cite{Lee2025} distinguishes between technological, environmental, and human factors. Other taxonomies differentiate between passive disengagements (system-detected limitations) and active, driver-initiated interventions~\cite{Lv2018}. Similarly,~\cite{Dixit2016} evaluates safety driver reaction times by classifying events into system failures, infrastructure issues, and external factors like weather or other road users. Recent analyses employ advanced computational methods to process these reports. Machine learning algorithms have been used for clustering and word-count analysis~\cite{Ward2024} and for predicting disengagement causes based on manufacturer, location, and initiator~\cite{Ali2025}. Also, Large Language Models (LLMs) were used to extract insights from the unstructured textual descriptions within these reports~\cite{Ansarinejad2025}.

Beyond \gls{cadmv} data, researchers explore disengagements through alternative methodologies. Some studies have analyzed human takeover reactions in Level~3 scenarios~\cite{Gershon2023}, while user interviews have informed conceptual frameworks for self-reported interventions~\cite{Nordhoff2024}. Simulated environments also offer safe testbeds for evaluating human factors and system behaviors, utilizing approaches such as analyzing safety driver brain activity for event classification ~\cite{Qi2025} or reinforcement learning to model disengagements~\cite{Zhou2025}.

A major limitation of studies analyzing \gls{cadmv} reports is that the evaluated software is proprietary, obscuring the algorithmic root causes of these disengagements. Deploying and evaluating open-source \gls{ad} stacks, such as Autoware Universe~\cite{Kato2018, AWrepo}, can help mitigate this opacity by enabling code-level inspection.

Although evaluating Autoware on small-scale mobile robots~\cite{Kim2024} or miniature vehicles to test digital twins~\cite{Samak2024} is well-documented, multiple studies highlight the complexities of deploying the stack on full-scale vehicles~\cite{Zhang2025, Ganesan2025, Korzelius2025}. Initial deployments often remain constrained to short, low-speed tests, e.g., parking-lot maneuvers~\cite{Zhang2025}, a \SI{100}{\meter} campus drive at up to~\SI{11}{\kilo\meter\per\hour}~\cite{Ganesan2025}, and a \SI{1.1}{\kilo\meter} campus route at up to~\SI{15}{\kilo\meter\per\hour}~\cite{Korzelius2025}.

Deployments in dynamic real-world traffic remain scarce. A notable exception is~\cite{Gu2024}, where Autoware was deployed on an autonomous bus on a public route in Fukaya, Japan. However, this study focused on validating a disengagement-logging tool rather than analyzing underlying causes. Another real-world Autoware deployment is discussed in~\cite{Liu2025Michigan}, though their safety-critical scenario evaluation relied on simulations and reported an estimated crash rate of $3\times10^{-3}$ per mile. 

Evaluating event severity requires robust criticality metrics, generally defined as the combined risk to involved actors~\cite{neurohrCriticalityAnalysisVerification2021}. Traditional conflict-centric measures like \gls{ttc}~\cite{haywardNearMissDeterminationUse1972} and \gls{pet} typically establish a \SI{1}{\second} near-miss threshold~\cite{haywardNearMissDeterminationUse1972, xuAnalyzingScenarioCriticality2024}. Because \gls{ttc} alone is often insufficient~\cite{linCommonRoadCriMeToolboxCriticality2023}, broader evaluation frameworks and toolboxes have emerged~\cite{schuttInverseUniversalTraffic2023, westhofenCriticalityMetricsAutomated2023}. Nevertheless, these metrics remain conflict-centric, making them ill-suited for non-collision system failures such as operational deadlocks or regulatory violations.

From the research gap derived above, we identify the following four key contributions:
\begin{itemize}
    \item A deployment and evaluation of an adapted Autoware Universe stack across \SI{236}{\kilo\meter} of mixed traffic in a representative real-world environment.
    \item A novel five-level criticality framework that classifies disengagements to distinguish between software limitations and over-cautious driver interventions.
    \item Identifying concrete system-level bottlenecks with clear ties to specific open-source modules.
    \item Providing a real-world benchmark for disengagement rate (\SI{0.127}{\per\kilo\meter}) for prototypical open-source stacks.
    
\end{itemize}


\section{Methodology}
\label{sec:method}
This section presents an analytical framework that defines disengagements for this study and categorizes them by criticality and root cause. This framework is later used for a detailed analysis of the observed disengagements, distinguishing between software limitations and human factors, and providing insights into the operational performance of the evaluated system.

\subsection{Definition of Disengagement}
\label{subsec:disengagement}

To enable precise root-cause attribution and encompass a broader spectrum of operational failures, this study adopts a functionally focused definition of disengagements from~\cite{CCRdisengagement}:\\
	\textbf{A disengagement from autonomous mode is an unplanned event in which a safety driver or remote operator assumes manual control to alter the vehicle's lateral or longitudinal behavior.}\\
This definition includes non-safety-critical operational failures, such as deadlocks, which stall the vehicle and expose system or module limitations. 
Under this definition, a disengagement can only occur while the vehicle is in autonomous mode, so startup problems, software or hardware issues before engagement, and planned exits are excluded.

\subsection{Disengagement Metrics}
To evaluate system performance, spatial and temporal disengagement rates~$\lambda_d$ and~$\lambda_t$ are defined as the number of disengagements $n$ normalized by total autonomous distance~$D$ and time~$T$ in \eqref{equation}.
\begin{equation}\label{equation}
    \lambda_d = \frac{n}{D} \quad \text{and} \quad \lambda_t = \frac{n}{T}.
\end{equation}
For an hourly analysis,~$\lambda_{T_i}$ represents the rate of interventions during hour~$T_i$ of the day across all rides, expressed as disengagements per~$60\,\unit{min}$ of driving time.

\subsection{Criticality Levels}
\label{subsec:definition_criticality}

\begin{table*}[t]
    \vspace{2mm}
    \centering
    \caption{Classification of Disengagement Criticality}
    \label{tab:criticality_levels}
    \renewcommand{\arraystretch}{1.2}%
    \begin{tabularx}{0.95\textwidth}{@{} c l X @{}}
        \toprule
        & \textbf{Criticality Category} & \textbf{Definition} \\
        \midrule
        \multirow{2}{*}{\blob{red}} & \multirow{2}{*}{\textbf{Critical Safety Failure}} & 
        Intervention was required to prevent an imminent collision, road departure, or significant safety hazard. Without this intervention, an accident was highly probable. \\
        
         \multirow{2}{*}{\blob{TUMWebOrange}} &  \multirow{2}{*}{\textbf{Precautionary (Recoverable)}} & 
        The safety driver intervened due to a perceived threat. Post hoc data analysis suggests the system was in a misbehaving state for a limited period, but would likely have recovered a safe state without the intervention. \\
        
         \multirow{2}{*}{\blob{TUMYellow}} &  \multirow{2}{*}{\textbf{Regulatory Violation}} & 
        Intervention required solely to prevent a violation of traffic regulations (e.g., traffic light violation) with no immediate physical danger to the vehicle or surroundings. \\

         \multirow{2}{*}{\blob{TUMBlueBrightDark}} &  \multirow{2}{*}{\textbf{Operational Deadlock}} & 
        The vehicle stopped safely but could not plan a valid trajectory to continue the mission (e.g., due to getting stuck behind obstacles), requiring manual intervention. \\
        
         \multirow{2}{*}{\blob{TUMWebGreen}} & \multirow{2}{*}{\textbf{False Positive Intervention}} & 
        The safety driver intervened due to subjective discomfort or caution. Subsequent data analysis confirms the system was operating within safe and legal bounds, rendering the intervention unnecessary. \\
        \bottomrule
    \end{tabularx}
\end{table*}

The occurrence of disengagements denotes only that autonomous mode was terminated, but it conveys no information about the underlying event type or its severity. As discussed in \autoref{sec:relatedwork}, existing criticality metrics focus primarily on conflict-centric scenarios~\cite{linCommonRoadCriMeToolboxCriticality2023} and are therefore unsuitable for non-conflict failures such as operational deadlocks or regulatory violations. To address this gap, we propose a five-level classification framework presented in~\autoref{tab:criticality_levels}, that accommodates both collision- and non-collision failure modes and supports an aggregate assessment of system performance.

The taxonomy encodes meta-information for assessing disengagement severity from both operator and developer perspectives. It explicitly includes a category for unnecessary safety-driver interventions and a precautionary category for interventions caused by confirmed, transient misbehavior that the system would likely have recovered from without manual input. Category assignment requires post hoc analysis of system logs and internal state.

\subsection{Classification of Root Causes}
For a detailed analysis of a disengagement, the root cause is as important as severity. Causes are organized by the modules of a modular \gls{ad} stack: Localization, Perception, Planning, and Control. Algorithmic failures and limitations are assigned to these categories. System failures can occur at the software infrastructure level (e.g., operating system or network issues). Hardware issues include failures of sensors and actuators, such as the vehicle actuation interface.\\
If a safety driver intervenes over-cautiously without any provable system failure, the root cause is external to the system and is assigned to the Human category.\\

\section{Experimental Setup \& Field Tests}
This section describes the experimental setup, on-road field tests, and the procedure for analyzing the collected data.
 
\subsection{Research Vehicle}
The test vehicle is a prototypical Level~4 research platform: a Volkswagen T7 Multivan (Fig.~\ref{fig:edgar}) adapted with an interface for software control of longitudinal and lateral actuators and equipped with multiple long- and short-range cameras and LiDAR sensors~\cite{EDGARpaper}.

The autonomous software is based on Autoware Universe~(v0.41.1)~\cite{AWrepo} and was extended via configuration and code changes for mixed-traffic driving. Enhancements include increasing the maximum speed to $100$~\si{\kilo\meter\per\hour}, handling dynamic electronic speed limits, retraining the LiDAR-only detector with additional data, and training a traffic-light classifier to detect green-arrow signals. Interfaces for remote vehicle operation were also integrated.~\cite{Teleoperation2022, ToD}.

\subsection{Route and Test Conditions}
\begin{figure*}[t]
    \vspace{2.0 mm}
    \centering
    \input{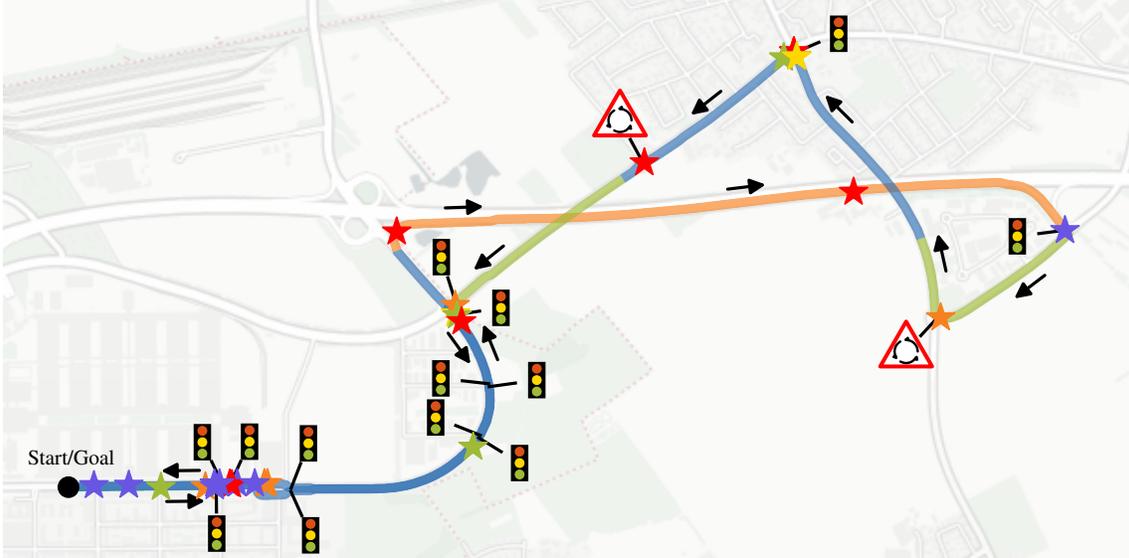}
    \vspace{-1.0 mm}
    \caption{Overview of the \SI{9.9}{\kilo\meter} suburban test route used for field evaluations. The route encompasses inner-city streets (blue), rural roads (green), and a highway segment (orange). Icons indicate the locations of traffic lights and roundabouts. Colored star markers denote the spatial distribution of the~30 recorded disengagement events, categorized by criticality as detailed in \autoref{tab:criticality_levels}. Map data: © OpenStreetMap contributors © CARTO.}
    \label{fig:map}
\end{figure*}

The testbed is a preselected suburban route representative of central Europe (\autoref{fig:map}). The route comprises inner-city streets (\SI{50}{\kilo\meter\per\hour}), rural roads (\SI{70}{\kilo\meter\per\hour}), and a highway segment with dynamic signage (\SIrange{80}{120}{\kilo\meter\per\hour}). Key route statistics are summarized in \autoref{tab:route_facts}.

\begin{table}[h]
    \centering
    \caption{Key Route Information.}
    \label{tab:route_facts}
    \begin{tabular}{lr}
        \toprule
        \textbf{Metric} & \textbf{Value} \\
        \midrule
        Max. length of one trip & \SI{9.9}{\kilo\meter} \\
        \quad thereof city streets & \SI{6.1}{\kilo\meter} \\
        \quad thereof rural streets & \SI{1.6}{\kilo\meter} \\
        \quad thereof highway & \SI{2.2}{\kilo\meter} \\
        \midrule
        Number of traffic lights & 13\\
        Number of unprotected left turns & 1 \\
        Number of roundabouts & 2 \\
        \bottomrule
    \end{tabular}
\end{table}

The selected route exposes the system to diverse operational conditions, multiple traffic lights, and both protected and unprotected turning maneuvers, creating a representative ODD. Unprotected left turns require gap-acceptance decisions, increasing demands on prediction and behavioral planning. Each ride included a safety driver able to assume manual control, a software operator responsible for starting and monitoring the \gls{ad} stack, and a remote operator capable of disengaging or guiding the vehicle~\cite{ToDpm,ToDeval}.
The route begins and ends at a private driveway where autonomous driving is prohibited, so the exact engagement timing varied from ride to ride.

\subsection{Data Collection and Analysis}
Data collection involved recording selected Autoware messages, sensor streams, and vehicle-state information, preserving the main inputs and outputs of each module.

All analyses were performed post hoc. Disengagement events were identified from the vehicle-interface state, and the preceding scene was reviewed by an expert annotator. For each event, the annotator assigned a criticality level, determined the causal chain, and identified the primary root cause. Following the definition in~\cite{Lv2018}, the results include only active disengagements, because the system cannot detect when it is operating outside its limitations.

\section{Results \& Discussion}
\label{sec:results}
The following presents quantitative results from the rides, followed by an analysis of disengagement types, their root causes, and representative case studies.

\subsection{Quantitative Results}
Across four consecutive days in September~2025, a total of~$26$ autonomous rides were conducted, covering \SI{236}{\kilo\meter}.\footnote{Note: Two additional rides were excluded from evaluation, as the LiDAR-based object detection was known to be nonfunctional.} Three rides were terminated before reaching the goal: two due to system failures that prevented continuation, and one due to a passenger's personal reason. 

\begin{figure*}[t]
    \vspace{1.0 mm}
    \centering
    %
    \begin{tikzpicture}[font=\footnotesize]
    \begin{groupplot}[
        group style={
            group size=1 by 3,
            vertical sep=0.15cm,
            x descriptions at=edge bottom
        },
        width=\textwidth,
        height=3.1cm,
        grid=major,
        grid style={dashed, gray!20},
        no markers,
        xmin=0, xmax=9.9,
        table/col sep=comma
    ]

        \nextgroupplot[clip=false, ylabel={$v$ in \si{\kilo\meter\per\hour}}, ymin=0, ymax=104]
            \begin{scope}[on background layer]
                \fill[TUMBlueBrand!15]   (axis cs:0,0)   rectangle (axis cs:2.4,104) node[black, midway, anchor=north, yshift=1.2cm, font=\scriptsize] {City};
                \fill[TUMWebOrange!15] (axis cs:2.4,0) rectangle (axis cs:4.6,104) node[black, midway, anchor=north, yshift=1.2cm, font=\scriptsize] {Highway};
                \fill[TUMWebGreen!15]  (axis cs:4.6,0) rectangle (axis cs:5.6,104) node[black, midway, anchor=north, yshift=1.2cm, font=\scriptsize] {Rural};
                \fill[TUMBlueBrand!15]   (axis cs:5.6,0) rectangle (axis cs:7.1,104) node[black, midway, anchor=north, yshift=1.2cm, font=\scriptsize] {City};
                \fill[TUMWebGreen!15]  (axis cs:7.1,0) rectangle (axis cs:7.7,104) node[black, midway, anchor=north, yshift=1.2cm, font=\scriptsize] {Rural};
                \fill[TUMBlueBrand!15]   (axis cs:7.7,0) rectangle (axis cs:9.9,104) node[black, midway, anchor=north, yshift=1.2cm, font=\scriptsize] {City};
            \end{scope}

            \foreach \i in {0,...,25} { 
                \addplot[TUMGrey3!35, thin] table [x=progress, y=ride_\i_v] {figures/timeseries_data.csv};
            }
            \addplot[TUMBlueBrand!100!black, very thick] table [x=progress, y=mean_v] {figures/timeseries_data.csv};

        \nextgroupplot[ylabel={$a_x$ in \si{\meter\per\square\second}}, ymin=-2.1, ymax=1.9]
            \begin{scope}[on background layer]
                \fill[TUMBlueBrand!15]   (axis cs:0,-2.1)   rectangle (axis cs:2.4,1.9);
                \fill[TUMWebOrange!15] (axis cs:2.4,-2.1) rectangle (axis cs:4.6,1.9);
                \fill[TUMWebGreen!15]  (axis cs:4.6,-2.1) rectangle (axis cs:5.6,1.9);
                \fill[TUMBlueBrand!15]   (axis cs:5.6,-2.1) rectangle (axis cs:7.1,1.9);
                \fill[TUMWebGreen!15]  (axis cs:7.1,-2.1) rectangle (axis cs:7.7,1.9);
                \fill[TUMBlueBrand!15]   (axis cs:7.7,-2.1) rectangle (axis cs:9.9,1.9);
            \end{scope}
            
            \foreach \i in {0,...,25} {
                \addplot[TUMGrey3!35, thin] table [x=progress, y=ride_\i_ax] {figures/timeseries_data.csv};
            }
            \addplot[TUMBlueBrand!100!black, very thick] table [x=progress, y=mean_ax] {figures/timeseries_data.csv};

        \nextgroupplot[ylabel={$a_y$ in \si{\meter\per\square\second}}, xlabel={Driven distance $s$ in \si{\kilo\meter}}, ymin=-2.2, ymax=1.8]
            \begin{scope}[on background layer]
                \fill[TUMBlueBrand!15]   (axis cs:0,-2.2)   rectangle (axis cs:2.4,1.9);
                \fill[TUMWebOrange!15] (axis cs:2.4,-2.2) rectangle (axis cs:4.6,1.9);
                \fill[TUMWebGreen!15]  (axis cs:4.6,-2.2) rectangle (axis cs:5.6,1.9);
                \fill[TUMBlueBrand!15]   (axis cs:5.6,-2.2) rectangle (axis cs:7.1,1.9);
                \fill[TUMWebGreen!15]  (axis cs:7.1,-2.2) rectangle (axis cs:7.7,1.9);
                \fill[TUMBlueBrand!15]   (axis cs:7.7,-2.2) rectangle (axis cs:9.9,1.9);
            \end{scope}

            \foreach \i in {0,...,25} {
                \addplot[TUMGrey3!35, thin] table [x=progress, y=ride_\i_ay] {figures/timeseries_data.csv};
            }
            \addplot[TUMBlueBrand!100!black, very thick] table [x=progress, y=mean_ay] {figures/timeseries_data.csv};

    \end{groupplot}
\end{tikzpicture}%

    \vspace{-2.0 mm}
    \caption{Vehicle dynamics across 26 autonomous rides. The panels display velocity $v$, longitudinal acceleration $a_x$, and lateral acceleration $a_y$ as a function of driven distance $s$. Individual rides are shown in gray, with the mean profile represented by the solid blue line. Background shading indicates the operational environment: city (blue), highway (orange), and rural (green).}
    \label{fig:spaghetti}
\end{figure*}
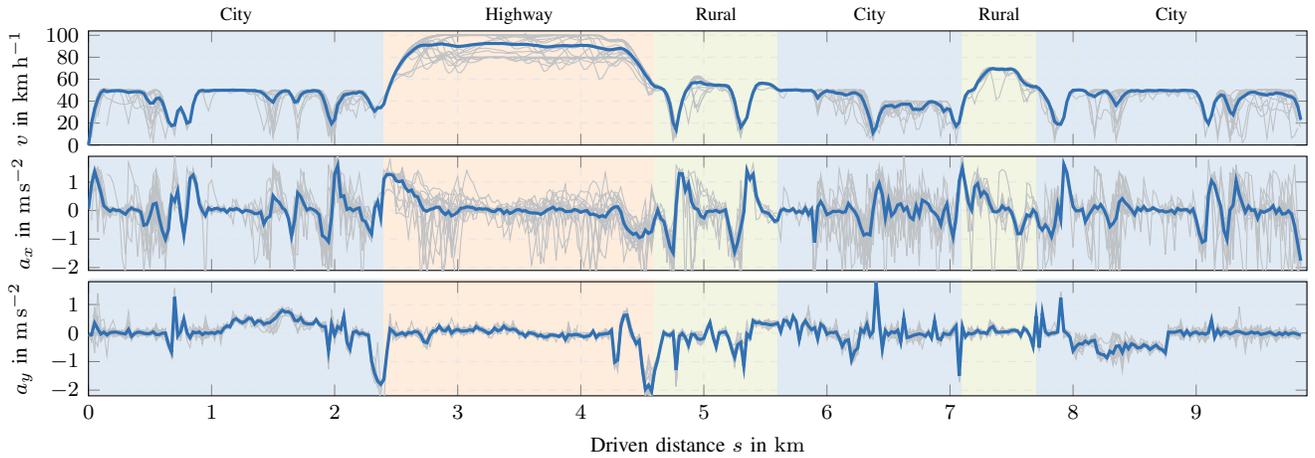

Vehicle velocity $v$, longitudinal acceleration~$a_x$, and lateral acceleration~$a_y$ across all rides are shown in \autoref{fig:spaghetti}. The velocity profile distinguishes the three street environments, with the highway segment showing the highest variance and typical speeds between \SI{80}{\kilo\meter\per\hour} and \SI{100}{\kilo\meter\per\hour}, while the acceleration profile shows a lot of variance in the braking behavior, due to varying traffic densities.
\begin{table}[bh]
    \centering
    \caption{Summary of the field study data and disengagement metrics.}
    \label{tab:field_study_results}
    \begin{tabular}{lr}
        \toprule
        \textbf{Metric} & \textbf{Value} \\
        \midrule
        Total number of trips & 26 \\
        Total distance of autonomous driving $D$ & \SI{236}{\kilo\meter} \\
        Total time of autonomous driving $T$ & \SI{378}{\min} \\
        Total number of disengagements $n$ & 30 \\
        Disengagements resolved by remote operator & 5 \\
        Number of rides aborted & 3 \\
        Number of rides without disengagement & 6 \\
        Highest number of disengagements per ride & 3 \\
        \midrule
        Mean disengagements per ride & 1.15 \\
        Spatial disengagement rate $\lambda_d$ & \SI{0.127}{\per\kilo\meter} \\
        Temporal disengagement rate $\lambda_t$ & \SI{4.76}{\per\hour} \\
        \bottomrule
    \end{tabular}
\end{table}
The general statistics are summarized in \autoref{tab:field_study_results}. The field tests resulted in $30$ disengagements across $26$ rides, yielding a spatial disengagement rate $\lambda_d$ of~\SI{0.127}{\per\kilo\meter}. Disengagements per ride ranged from~$0$ to~$3$, and five disengagements were initiated by a remote operator to resolve a deadlock situation. Relative to simulation-based evaluations~\cite{Liu2025Michigan} and established commercial operators~\cite{Kohanpour2025}, the observed disengagement rate is substantially higher, demonstrating both the sim-to-real gap and performance differences between proprietary and open-source stacks. This study thus establishes a necessary benchmark for real-world open-source systems.

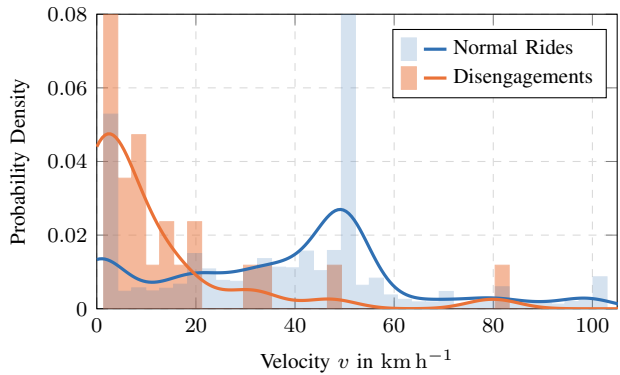
\begin{figure}[h]
    \centering
    %
    \begin{tikzpicture}[font=\footnotesize]
        \begin{axis}[
            width=8.5cm,
            height=5.5cm,
            tick label style={/pgf/number format/fixed},
            scaled ticks=false,
            xlabel={Velocity $v$ in \si{\kilo\meter\per\hour}},
            ylabel={Probability Density},
            ymin=0, ymax=0.08,
            xmin=0, xmax=105,
            grid=major,
            grid style={dashed, gray!30},
            legend style={at={(0.98,0.95)}, anchor=north east},
            legend cell align={left},
            table/col sep=comma
        ]            
            \addplot[ybar interval, fill=TUMRed!100, draw=TUMRed!100, opacity=0.5, forget plot] 
                table[x=bin_x, y=dis_density] {figures/vel_pdf_hist_data.csv};
            
            \addplot[ybar interval, forget plot, fill=TUMBlueBrand!100, draw=TUMBlueBrand!100, opacity=0.2] 
                table[x=bin_x, y=pop_density] {figures/vel_pdf_hist_data.csv};

            \addplot[TUMBlueBrand!100!black, very thick, no markers, forget plot] 
                table[x=x, y=pop_kde] {figures/vel_pdf_kde_data.csv};

            \addlegendimage{
            legend image code/.code={
                \draw[fill=TUMBlueBrand!100, opacity=0.2, draw=TUMBlueBrand!100] (0cm,-0.15cm) rectangle (0.2cm,0.15cm);
                \draw[TUMBlueBrand!100!black, very thick] 
                    (0.3cm,0cm) -- (0.6cm,0cm);
                }
            }
            \addlegendentry{Normal Rides}

            \addplot[TUMRed!100!black, very thick, no markers] 
                table[x=x, y=dis_kde, forget plot] {figures/vel_pdf_kde_data.csv};
            \addlegendimage{
            legend image code/.code={
                \draw[fill=TUMRed!100, opacity=0.5, draw=TUMRed!100] (0cm,-0.15cm) rectangle (0.2cm,0.15cm);
                \draw[TUMRed!100!black, very thick] 
                    (0.3cm,0cm) -- (0.6cm,0cm);
                }
            }
            \addlegendentry{Disengagements}
        
        \end{axis}
\end{tikzpicture}%

    \vspace{-2.0 mm}
    \caption{Histogram plots and Kernel Density Estimates (KDE)~\cite{Scott2015} for the velocities of all timestamps for all rides (blue) and the velocities at the moment of the disengagements (red).}
    \label{fig:vel_histo}
\end{figure}

The velocity distributions over all timestamps (blue) and at disengagement (red) are compared in \autoref{fig:vel_histo}. A two-sample Kolmogorov-Smirnov test~\cite{Hodges1958} confirms a significant difference between the distributions ($p=1.73\times10^{-11}$), indicating that disengagements occur more frequently at lower speeds, which is consistent with the occurrence of scenarios involving static objects or traffic lights.

\begin{figure}[h]
    \centering
    %
    \begin{tikzpicture}[font=\footnotesize]
    \begin{axis}[
        width=0.5\textwidth,
        height=5.5cm,
        ybar,
        bar width=0.8cm,
        xlabel={Time of Day in \si{\hour}},
        ylabel={$\lambda_{T_i}$ in \si{\per\hour}},
        xmin=8.25, xmax=16.75,
        ymin=0, ymax=13.5,
        xtick={9,10,11,12,13,14,15,16},
        grid=major,
        xmajorgrids=false,
        ymajorgrids=true,
        grid style={dashed, gray!30},
        point meta=explicit, 
        nodes near coords={
            \begin{tabular}{c}
                \pgfmathprintnumber[precision=0]{\pgfplotspointmeta}\, dis.\\
                \pgfmathprintnumber[fixed, precision=0]{\mins}\,min
            \end{tabular}
        },
        every node near coord/.append style={
            font=\scriptsize,
            yshift=2pt,
            align=center,
            black
        },
        tick style={draw=none}
    ]
        \addplot[
            fill=TUMBlueBrand!60, 
            draw=TUMBlueBrand!100,
            visualization depends on={value \thisrow{minutes} \as \mins},
        ] 
        table [
            x=hour, 
            y=rate, 
            meta=events, 
            col sep=comma
        ] {figures/hourly_rates_data.csv};
        
    \end{axis}
\end{tikzpicture}%

    \vspace{-6.0 mm}
    \caption{Temporal disengagement rate $\lambda_{T_i}$ for the hours of a day. The bars show the disengagement rate for a specific hour, while the annotations show the number of disengagements and the total number of minutes driven autonomously within these hours, across all rides over multiple days.}
    \label{fig:hourly_rates}
\end{figure}
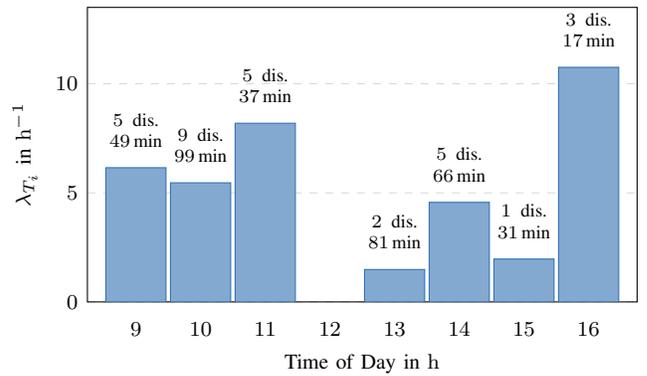

\autoref{fig:hourly_rates} shows the hourly disengagement rates. A two-sample Kolmogorov-Smirnov test~\cite{Hodges1958} comparing minutes driven per hour with disengagements per hour shows no significant dependence on time of day ($p=0.529$), although the limited sample size and exposure across hours should be considered when interpreting this result.

\subsection{Case Study of Disengagement Scenarios}
To understand the underlying causes and failure modes, a detailed analysis of disengagements across the different root cause categories is presented. \autoref{fig:map} shows the locations of all disengagements, color-coded by their criticality level.

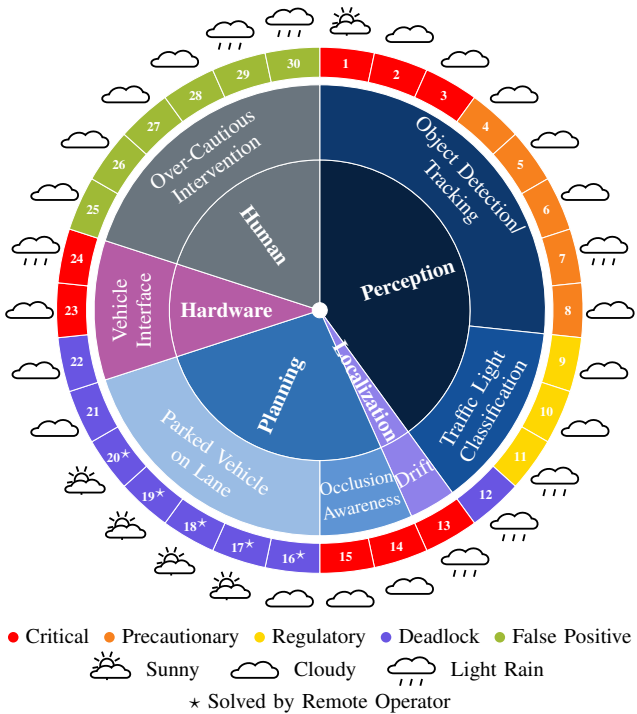
\begin{figure}[h]
    \centering
    %
\newcommand{\wxSun}{\tikz[baseline=-0.6ex, scale=0.15]{
  \draw[line width=0.6pt] (0,0) circle (0.9);
  \foreach \a in {0,45,...,315}{
    \draw[line width=0.6pt] (\a:1.2) -- (\a:1.9);
  }
}}

\newcommand{\wxSunCloud}{%
\tikz[baseline=-0.6ex, scale=0.15]{
  \draw[line width=0.7pt] (-0.5,1.1) circle (0.6);
  \foreach \a in {30,70,...,330}{
    \draw[line width=0.7pt]
      (-0.5,1.1) ++(\a:0.8) -- ++(\a:0.5);
  }

  \fill[white]
    (-1.5,0) .. controls (-2.1,0) and (-2.1,0.6) .. (-1.4,0.6)
    .. controls (-1.2,1.1) and (-0.4,1.2) .. (-0.1,0.9)
    .. controls (0.2,1.3) and (0.9,1.2) .. (1.1,0.7)
    .. controls (1.7,0.7) and (1.7,0) .. (1.2,0)
    -- cycle;

  \draw[line width=0.7pt]
    (-1.5,0) .. controls (-2.1,0) and (-2.1,0.6) .. (-1.4,0.6)
    .. controls (-1.2,1.1) and (-0.4,1.2) .. (-0.1,0.9)
    .. controls (0.2,1.3) and (0.9,1.2) .. (1.1,0.7)
    .. controls (1.7,0.7) and (1.7,0) .. (1.2,0)
    -- cycle;
}}

\newcommand{\wxCloudy}{%
\tikz[baseline=-0.6ex, scale=0.15]{
  \draw[line width=0.7pt]
    (-1.6,0) .. controls (-2.3,0) and (-2.3,0.7) .. (-1.5,0.7)
    .. controls (-1.3,1.3) and (-0.4,1.5) .. (0,1.1)
    .. controls (0.4,1.6) and (1.3,1.5) .. (1.5,0.9)
    .. controls (2.2,0.9) and (2.2,0) .. (1.6,0)
    -- cycle;
}}

\newcommand{\wxCloudRain}{%
\tikz[baseline=-0.6ex, scale=0.15]{
  \draw[line width=0.7pt]
    (-1.6,0.4) .. controls (-2.3,0.4) and (-2.3,1.1) .. (-1.5,1.1)
    .. controls (-1.3,1.7) and (-0.4,1.9) .. (0,1.5)
    .. controls (0.4,2.0) and (1.3,1.9) .. (1.5,1.3)
    .. controls (2.2,1.3) and (2.2,0.4) .. (1.6,0.4)
    -- cycle;

  \foreach \x in {-0.9,0,0.9}{
    \draw[line width=0.7pt] (\x,-0.1) -- ++(-0.15,-0.6);
  }
}}

\newcommand{\ToDstar}{\raisebox{0.65ex}{\scriptsize$\star$}}

\begin{tikzpicture}

\wheelchart[
    middle style={darkgray, font=\bfseries\large},
    radius={0.1}{2},
    slices style={fill=\WCvarB, draw=white},
    data=,
    wheel data=\WCvarC,
    wheel data pos=0.6,
    wheel data style={white, font=\footnotesize\bfseries, rotate={(\WCmidangle>90 && \WCmidangle<270) ? \WCmidangle+180 : \WCmidangle}}
]{%
    12/TUMBlueDark/Perception,
    1/TUMBlueBright/Localization,
    8/TUMBlueBrand/Planning,
    3/TUMPink/Hardware,
    6/TUMGrey4/Human
}

\wheelchart[
    radius={2}{3},
    slices style={fill=\WCvarB, draw=white},
    data=,
    wheel data=\WCvarC,
    wheel data pos=0.5,
    wheel data style={white, font=\footnotesize,
    align=center, rotate={(\WCmidangle>180 && \WCmidangle<360) ? \WCmidangle+90 : \WCmidangle-90}},
    start angle=90
]{%
    8/TUMBlueDark2/Object Detection{/}\\ Tracking, 
    4/TUMBlueDark4/Traffic Light \\ Classification,
    1/TUMBlueBright/Drift,
    2/TUMBlueLight/\scriptsize Occlusion\\\scriptsize Awareness,
    6/TUMBlueLightDark/ Parked Vehicle\\ on Lane,
    3/TUMPink/Vehicle\\ Interface,
    6/TUMGrey4/Over-Cautious\\ Intervention
}

\wheelchart[
    radius={3.1}{3.5}, 
    slices style={fill=\WCvarB, draw=white},
    data=,
    wheel data=\WCvarD,           
    wheel data pos=0.5,           
    wheel data style={white, font=\tiny\bfseries},
    start angle=90
]{%
    1/red/Critical/{1},
    1/red/Critical/{2},
    1/red/Critical/{3},
    1/TUMWebOrange/Precautionary/{4},
    1/TUMWebOrange/Precautionary/{5},
    1/TUMWebOrange/Precautionary/{6},
    1/TUMWebOrange/Precautionary/{7},
    1/TUMWebOrange/Precautionary/{8},
    1/TUMYellow/Regulatory/{9},
    1/TUMYellow/Regulatory/{10},
    1/TUMYellow/Regulatory/{11},
    1/TUMBlueBrightDark/Regulatory/{12},
    1/red/Critical/{13},
    1/red/Critical/{14},
    1/red/Critical/{15},
    1/TUMBlueBrightDark/Deadlock/{16\ToDstar},
    1/TUMBlueBrightDark/Deadlock/{17\ToDstar},
    1/TUMBlueBrightDark/Deadlock/{18\ToDstar},
    1/TUMBlueBrightDark/Deadlock/{19\ToDstar},
    1/TUMBlueBrightDark/Deadlock/{20\ToDstar},
    1/TUMBlueBrightDark/Deadlock/{21},
    1/TUMBlueBrightDark/Deadlock/{22},
    1/red/Critical/{23},
    1/red/Critical/{24},
    1/TUMWebGreen/Not needed/{25},
    1/TUMWebGreen/Not needed/{26},
    1/TUMWebGreen/Not needed/{27},
    1/TUMWebGreen/Not needed/{28},
    1/TUMWebGreen/Not needed/{29},
    1/TUMWebGreen/Not needed/{30}
}

\wheelchart[
    radius={3.75}{3.95},
    slices style={fill=white, draw=white}, 
    data=,
    wheel data=\WCvarD,
    wheel data pos=0.55,
    wheel data style={font=\tiny},
    start angle=90,
]{%
    1/white//{\wxSunCloud},
    1/white//{\wxCloudy},
    1/white//{\wxCloudy},
    1/white//{\wxCloudy},
    1/white//{\wxCloudy},
    1/white//{\wxCloudy},
    1/white//{\wxCloudRain},
    1/white//{\wxCloudy},
    1/white//{\wxCloudy},
    1/white//{\wxCloudy},
    1/white//{\wxCloudRain},
    1/white//{\wxCloudRain},
    1/white//{\wxCloudRain},
    1/white//{\wxCloudy},
    1/white//{\wxCloudy},
    1/white//{\wxCloudy},
    1/white//{\wxSunCloud},
    1/white//{\wxSunCloud},
    1/white//{\wxSunCloud},
    1/white//{\wxSunCloud},
    1/white//{\wxCloudy},
    1/white//{\wxCloudy},
    1/white//{\wxCloudy},
    1/white//{\wxCloudRain},
    1/white//{\wxCloudy},
    1/white//{\wxCloudy},
    1/white//{\wxCloudy},
    1/white//{\wxCloudy},
    1/white//{\wxCloudRain},
    1/white//{\wxCloudRain}
}

\node[anchor=north, yshift=-0.0cm] (leg1) at (current bounding box.south) {
    \footnotesize
    \tikz\fill[red] (0,0) circle (0.07); Critical \ 
    \tikz\fill[TUMWebOrange] (0,0) circle (0.07); Precautionary \
    \tikz\fill[TUMYellow] (0,0) circle (0.07); Regulatory \
    \tikz\fill[TUMBlueBrightDark] (0,0) circle (0.07); Deadlock \
    \tikz\fill[TUMWebGreen] (0,0) circle (0.07); False Positive 
    
};
\node[anchor=north, yshift=0.2cm] (leg2) at (leg1.south) {
    \footnotesize

    \raisebox{-1mm}{\wxSunCloud}\; Sunny \quad
    \raisebox{-1mm}{\wxCloudy}\; Cloudy \quad
    \raisebox{-1mm}{\wxCloudRain}\; Light Rain
    \par\vspace{0.2em}
};
\node[anchor=north, yshift=0.15cm] (leg3) at (leg2.south) {
    \footnotesize
    $\star$ Solved by Remote Operator
};
\end{tikzpicture}%

    \vspace{-6.0 mm}
    \caption{Classification of all disengagements by root cause, criticality, and weather conditions. The inner ring identifies the primary root cause for the intervention. The middle ring further breaks these down into specific failure modes. The outer numbered ring color-codes each event by its criticality level as defined in \autoref{tab:criticality_levels}. Surrounding icons indicate the weather conditions at the time of each event, following the taxonomy in~\cite{nuScenes}, where Sunny indicates direct sunlight and Light Rain indicates sensor contamination.}
    \label{fig:piechart}
\end{figure}

\autoref{fig:piechart} presents a hierarchical classification of disengagements by root cause and criticality level. Analysis of the $26$ rides reveals that $40\,\%$ of disengagements originated from perception-related failures, followed by $26.7\,\%$ from limitations in the planning module. While these proportions roughly align with broader field-test trends~\cite{Boggs2020, Ward2024, Kohanpour2025}, precise comparisons remain challenging due to the proprietary nature of \gls{ad} software in the existing literature.

\subsection*{Hardware and Localization Failures}
Three disengagements occurred due to hardware interface failures, including a software-to-vehicle interface disconnect. In two instances, disconnection occurred during active steering maneuvers, resulting in immediate loss of vehicle control and near-collision with road boundaries. The third disconnection occurred while stationary at a traffic light, but proved irreversible without a complete system restart, an operation infeasible during traffic, resulting in an operational deadlock requiring manual intervention.

A separate critical failure occurred in the localization module, where the pose estimate drifted approximately~\SI{2}{\meter} while the vehicle navigated a highway segment. This offset persisted for the remainder of the ride, rendering safe re-engagement impossible and necessitating termination of the ride. Root cause analysis could not definitively distinguish between erroneous sensor measurements or incorrect sensor data processing, but the single-occurrence nature suggests GNSS measurement failure.

\subsection*{Perception Failures}
A subset of perception-related failures involved misclassification of traffic signal states. Four disengagements resulted from traffic light classifier errors: three instances of red signals misclassified as green or off (regulatory violations) and one instance of a green signal failing to be detected (operational deadlock).

Analysis across all autonomous rides reveals that $312$ out of $316$ traffic lights were passed without disengagement, yielding an entity-based success rate of $98.7\,\%$. Because the traffic light logic requires only a few consecutive misclassified frames to trigger a disengagement, the frame-based error rate on this route is likely below $1\,\%$. While this would typically be considered sufficient for many learning-based applications, in the \gls{ad} domain, these error rates lead to multiple disengagements over realistic driving distances. This finding reinforces the necessity of real-world validation in the specific application domain rather than relying solely on dataset-derived performance metrics.

The most frequently occurring perception failures involved loss of previously detected objects. Eight disengagements occurred when previously tracked objects became undetected despite remaining physically present. Fig.\ref{fig:perception_scenario} illustrates a representative example in which the perception system lost a bus positioned ahead of the ego vehicle, prompting the planner to execute forward acceleration into an occupied space. This pattern of anomalies in the traffic light and object perception suggests a robustness limitation within the \gls{ad} stack under real-world conditions.

\subsection*{Planning Failures}
\begin{figure}[b]
    \vspace{-2.0 mm}
    %
    \input{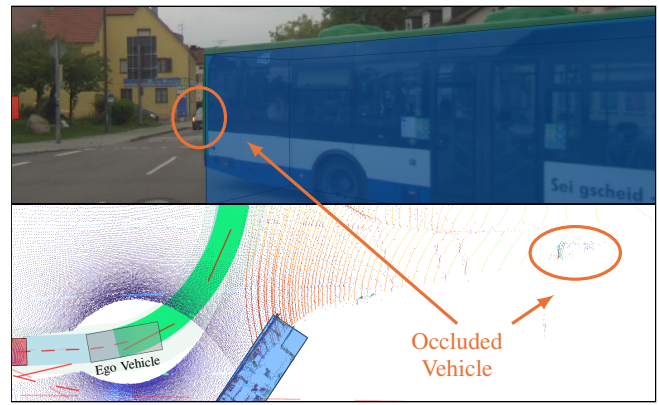}%

    \vspace{-4mm}
    \caption{The ego vehicle performs an unprotected left turn (green trajectory). An oncoming vehicle, which had right of way, was occluded by a bus and detected too late for the system to react, so the safety driver intervened.}
    \label{fig:occlusion_scenario}
\end{figure}

Two planning-related disengagements resulted from hazardous vehicle interactions involving occlusion by other traffic participants. \autoref{fig:occlusion_scenario} illustrates a representative scenario: an oncoming vehicle in the opposing traffic lane, partially occluded by an interposing bus, was detected too late for the ego vehicle to brake sufficiently while maintaining right-of-way compliance and collision avoidance. Although the perception system provided correct output under the occlusion constraint, the planning module lacks mechanisms to anticipate threats obscured by occlusion.

Autoware's prototypical \textit{Occlusion Spot Module} addresses only pedestrian occlusions, and the \textit{Intersection Module's} capabilities proved insufficient for our scenarios. Therefore, these modules were deactivated. Thus, occlusion-aware planning should consider all road users, not only pedestrians. Although algorithmic solutions exist~\cite{Occlusion1, Occlusion2, Moller2025}, real-world behavioral-planning implementations in open-source stacks like Autoware are needed.

\begin{figure}[h]
    %
    \input{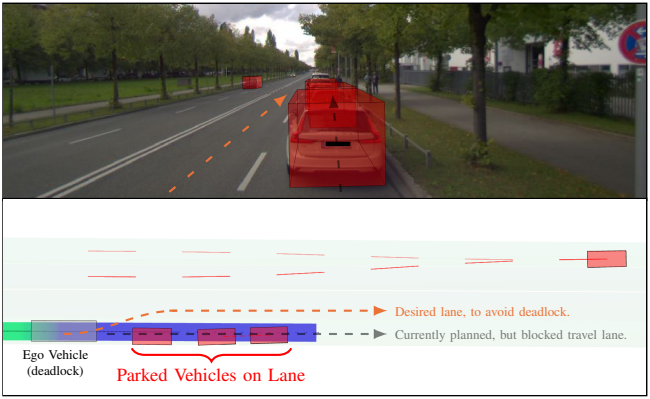}%

    \vspace{-4mm}
    \caption{Representative operational deadlock scenario caused by planning module limitations. The ego vehicle becomes stuck behind stationary vehicles parked in the travel lane, as shown by the red bounding boxes. While the desired behavior is to circumnavigate the obstacles, the current planning stack remains on the blocked travel lane.}
    \vspace{-2mm}
    \label{fig:parking_awareness}
\end{figure}

Six planning-related disengagements occurred in the final segment when stationary vehicles occupied the travel lane (\autoref{fig:parking_awareness}). The \textit{Static Obstacle Avoidance Module}, designed to circumnavigate lane-edge obstacles, cannot reliably decide whether to avoid stationary vehicles in the travel lane. This reflects an operational risk: circumnavigation attempts are often counterproductive, e.g., when a vehicle waits in a queue behind a stopped vehicle at a traffic light. The planning module lacks situational context to distinguish short-term waiting queues from long-term parking, creating deadlocks.

Resolution of this limitation would require assessing situational awareness to classify vehicle intent. Potential indicators include hazard light status, brake light illumination, delivery-related markings, or pedestrian activity near the vehicle. However, implementing rule-based classification schemes for these indicators presents substantial complexity. More robust solutions require models with broader semantic reasoning capabilities to support contextual decision-making in ambiguous scenarios.

\subsection*{Human-caused Disengagements}
Six disengagements were initiated by the safety driver despite the autonomous system operating within safe and legal boundaries. Post hoc analysis confirmed that the system would have likely resolved these situations, rendering the interventions unnecessary. These events reveal a systems-level weakness distinct from software or hardware failures: the safety driver's confidence in the system's reliability did not align with the system's actual capability.

The observed interventions primarily involved two scenarios: five instances during approaches toward detected objects, likely related to driver concern about perception reliability, and one instance at a traffic signal stop line. This pattern indicates insufficient feedback mechanisms for conveying the system's confidence to the safety driver and a lack of trust on the safety driver's part. Reducing this kind of interventions reduces the total number of disengagements and helps building trust in autonomous systems.

\subsection{Discussion}
This field study and the accompanying disengagement analysis provide practical insights into the reproducibility and failure modes of an open-source modular \gls{ad} stack. However, several limitations should be emphasized.

First, the analysis was performed post hoc, using only the recorded data. While this reduces bias from supplementary commentary, it omits contextual information that safety drivers or remote operators could provide, such as perceived intent or environmental cues. Future studies could augment recorded data with structured driver debriefs, short-form annotations recorded at the time of a disengagement, or audio context to improve root-cause attribution.

Second, the criticality taxonomy, as defined in \autoref{tab:criticality_levels}, captures annotator-perceived severity but remains subjective. To increase labeling reliability, it could be valuable to supplement human labels with objective measures such as \gls{ttc}, \gls{pet}, or other surrogate-safety metrics or with composite scores for conflict-centered scenarios. Other disengagement types, such as deadlocks, regulatory violations, and false positives, are inherently difficult to reduce to a single numeric criticality, so a mixed qualitative–quantitative approach is preferable.

Third, the fixed-route highlights reproducibility constraints but limits external validity. The results represent some mixed-traffic environments, but broader generalization requires additional routes, greater traffic, and environmental condition variance, and a larger sample of rides.

Finally, several operational insights point to actionable improvements: better conveying system confidence to safety drivers to reduce unnecessary interventions, implementing occlusion- and situation-aware planning strategies, and improving perception robustness for rare but safety-relevant cases, such as traffic-light misclassifications. Addressing these failure modes could improve performance and safety, reduce the observed disengagement rate, and enable more reliable, scalable real-world deployments.

Taken together, these limitations do not undermine the study's primary contributions, but they highlight directions for increasing the rigor and generality of future field evaluations.

\section{Conclusion \& Outlook}
\label{sec:conclusion}

This field study analyzed disengagements of an open-source autonomous driving stack across~\SI{236}{\kilo\meter} of mixed-traffic operation.~$30$~disengagements (\SI{0.127}{\per\kilo\meter}) were recorded and evaluated using a five-level criticality framework that accommodates both collision and non-collision failures. Root causes stemmed predominantly from perception failures~(\SI{40}{\percent}), planning limitations~(\SI{26.7}{\percent}), and over-cautious safety drivers~(20\,\%).\\
Perception emerged as a critical bottleneck: rare tracking losses and traffic light misclassifications necessitated safety-relevant interventions despite entity-based success rates exceeding 98\,\%. This discrepancy highlights how infrequent anomalies, when aggregated, can lead to system-level failures over extended operation, underscoring the insufficiency of offline evaluation. Furthermore, planning limitations exposed inadequate situational awareness, particularly regarding occlusion-aware threat assessment and the handling of stationary obstacles blocking full lanes. These findings are specific to our fixed-route ODD and limited exposure, which constrains generalization.\\
Future work should expand empirical evaluations across diverse conditions and ODDs, incorporating criticality frameworks and multiple real-world performance metrics.


\section*{Acknowledgement}
\noindent We would like to thank D. Kulmer, I. Tahiraj, C. Schröder, D. Brecht, E. Rivera, I. Trautmannsheimer, L. Stratil, N. Karunainayagam, N. Krauss, N. Gehrke, R. Taupitz, R. Stolz, T. Kerbl, T. Mascetta, and X. Su for their support in collecting the data.


\bibliographystyle{IEEEtran}
\bibliography{literature}
\end{document}